\documentclass[conference]{IEEEtran}
\usepackage{tikz}
\usetikzlibrary{shapes.geometric, arrows, positioning}
\usepackage{amsmath,amssymb,amsfonts}
\usepackage{balance} 
\usepackage{booktabs} 
\usepackage{multirow}
\usepackage{algpseudocode}
\usepackage[greek,english]{babel}
\usepackage{lipsum}
\usepackage{mathtools}
\usepackage{cuted}
\usepackage{textcomp}

\usepackage{gensymb}
\usepackage{amsmath} 
\usepackage{booktabs}
\usepackage{algorithm}
\usepackage{algpseudocode}
\usepackage[normalem]{ulem}
\usepackage[justification=centering]{caption} 
\usepackage{subcaption}
 \usepackage{lipsum}
\usepackage{hyperref} 
\usepackage{fullpage}
 \usepackage{xargs}                      

\def\BibTeX{{\rm B\kern-.05em{\sc i\kern-.025em b}\kern-.08em
    T\kern-.1667em\lower.7ex\hbox{E}\kern-.125emX}}
\tikzstyle{process} = [rectangle, rounded corners, minimum width=3cm, minimum height=1cm, text centered, draw=black, fill=teal!50]
\tikzstyle{data} = [cylinder, shape border rotate=90, minimum height=2cm, minimum width=1.5cm, text centered, draw=black, fill=white!50, aspect=0.75]
\tikzstyle{decision} = [rectangle, rounded corners, minimum width=3cm, minimum height=1cm, text centered, draw=black, fill=yellow!50]
\tikzstyle{arrow} = [thick,->,>=stealth]

\expandafter\def\expandafter\normalsize\expandafter{%
    \normalsize%
    \setlength\abovedisplayskip{0pt}%
    \setlength\belowdisplayskip{8pt}%
    \setlength\abovedisplayshortskip{-8pt}%
    \setlength\belowdisplayshortskip{2pt}%
}
\title{ Off-Policy Evaluation and Counterfactual Methods   in Dynamic Auction Environments}

 \author{ \IEEEauthorblockN{Ritam Guha\IEEEauthorrefmark{1}  and Nilavra Pathak\IEEEauthorrefmark{2}   } \IEEEauthorblockA{\IEEEauthorrefmark{1}Department of Computer Science and Engineering, Michigan State University, East Lansing, MI}   
  \IEEEauthorblockA{\IEEEauthorrefmark{2} Marketing Data Science, Expedia Group,  New York, NY}  
 \IEEEauthorblockA{guharita@msu.edu, npathak@expediagroup.com }}

\begin{document}

\maketitle
\begin{abstract}  
\sloppy
Counterfactual estimators are critical for learning and refining policies using logged data, a process known as Off-Policy Evaluation (OPE). OPE allows researchers to assess new policies without costly experiments, speeding up the evaluation process. Online experimental methods, such as A/B tests, are effective but often slow, thus delaying the policy selection and optimization process.

In this work, we explore the application of OPE methods in the context of resource allocation in dynamic auction environments. Given the competitive nature of environments where rapid decision-making is crucial for gaining a competitive edge, the ability to quickly and accurately assess algorithmic performance is essential. By utilizing counterfactual estimators as a preliminary step before conducting A/B tests, we aim to streamline the evaluation process, reduce the time and resources required for experimentation, and enhance confidence in the chosen policies. Our investigation focuses on the feasibility and effectiveness of using these estimators to predict the outcomes of potential resource allocation strategies, evaluate their performance, and facilitate more informed decision-making in policy selection. Motivated by the outcomes of our initial study, we envision an advanced analytics system designed to seamlessly and dynamically assess new resource allocation strategies and policies.

\end{abstract}

\section{Introduction}

Modern-day marketplaces are increasingly designed as two-sided platforms. In such marketplaces, the  goal is to maximize the utility on both  the buyers and the sellers. This in turn enhances the overall revenue generated within the marketplace, and is critical in ensuring the long-term viability and competitiveness of the platform. Marketplaces typically implement complex auction mechanisms to achieve these goals. Unlike traditional auctions, that focus solely on price-based allocation, modern auctions consider a variety of factors. It includes the strategic behaviors of both buyers and sellers, the dynamic nature of their payment policies, and the overall impact on marketplace revenue~\cite{milgrom2021auction, haeringer2018market}.  Key industrial applications of these advanced auction mechanisms include Energy Markets, Telecommunications, Digital Marketing and many others.

Given the complexity of these auctions, payment policies have also become increasingly dynamic~\cite{haeringer2018market}. Validating a new payment policy in the real-world settings presents significant challenges. A/B testing is commonly used for online evaluation of the new policies, which is both expensive and time-consuming. Moreover, in user-facing commercial applications, A/B tests can risk exposing users to poor-performing policies, which can directly harm the application's revenue. As an alternative, simulation-based learning has been proposed as a new direction for evaluation~\cite{jeunen2022learning}. However, these systems are still in their infancy, often relying on a deep understanding of the auction environment or on assumptions that may not hold true in real-world scenarios. While offline validation has its drawbacks, particularly the necessity of logged data, it has demonstrated significant advantages in applications such as recommendation and ranking. This method, known as Off-Policy Evaluation (OPE)~\cite{saito2020open}, is employed in reinforcement learning and decision-making contexts to estimate the performance of a new policy using historical data collected under a different policy. Essentially, OPE predicts how well a new or hypothetical strategy would perform based on data generated by an existing strategy. In industrial settings, particularly in recommendation and ranking systems, OPE provides a practical and efficient alternative to traditional online evaluation methods.

In this paper, we explore the limitations and potential of using OPE in a dynamic auction environment. Our goal is to develop a measurement system capable of evaluating the payment policies using existing data alone, thereby enabling us to assess the feasibility of improving performance by testing new policies. To evaluate the viability of this approach, our study addresses the following key research questions:

\begin{itemize}
   \item Conduct a comprehensive analysis of discrete versus continuous policy evaluation methods for dynamic competitive auctions.
   \item Assess the feasibility of identifying the best-performing policy among three options based on the outcomes of two prior tests.
   \item  Learning optimal policies through optimization of policy estimators and evaluating their effectiveness via simulation. 
\end{itemize}

\section{Overview of the Approach}

\subsection{Problem Description}

In this dynamic auction environment, resources are considered to be theoretically infinite, and an indeterminate number of agents compete to acquire them. Each agent assigns a personal value to the resources and aims to maximize their utility through participation. Agents operate according to a payment policy ($\pi$), which influences their willingness to pay to acquire the resources for which they are competing. 

The determination of the winning agent involves considering both the payment amount and the potential revenue generated from the allocation~\cite{haeringer2018market}. Winning the auction incurs costs for the agent, related to the resources expended. The effectiveness of the allocation is evaluated through metrics such as \textit{reach} , \textit{resources}, and \textit{returns} . \textit{Reach} measures the impact or extent of the allocation, this is analogous to views in marketing. \textit{Resources} denote  the quantity of units allocated after conversion, such as purchases, download, etc. \textit{Returns} reflect the overall utility of winning the allocation.

Agents have the flexibility to design their payment policies to optimize their performance based on these metrics. This allows them to balance the costs incurred with the expected benefits derived from the allocated resources.

\subsection{Evaluation Pipeline}

\begin{figure}[htbp]
\centerline{\includegraphics[width=0.45\textwidth]{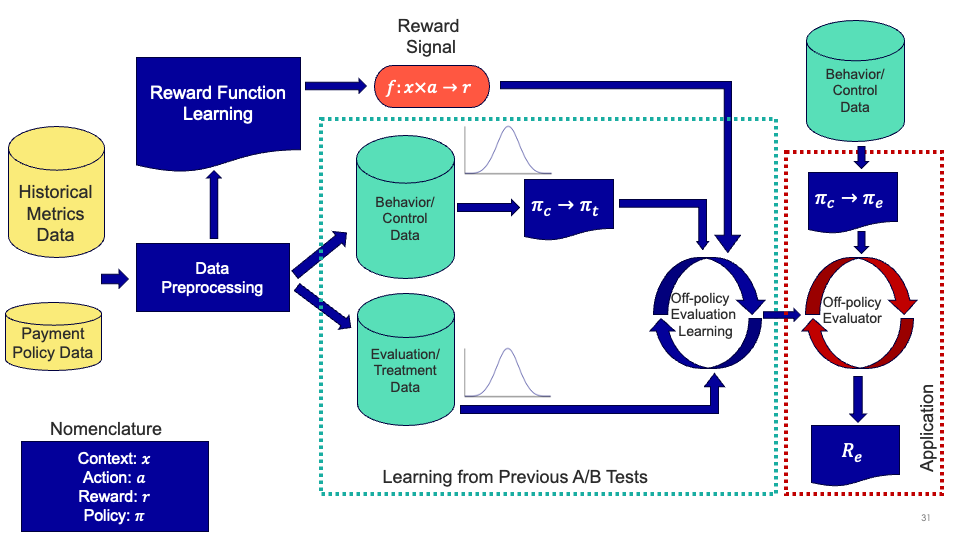}}
\caption{Off-Policy Evaluator Hyperparameter Learning and Application.}
\label{fig:overall}
\end{figure}

We adopt our pipeline based on the Open-Bandit Modules~\cite{saito2020open} but adapt it for our  use-cases. The only controllable units in this case are the payment policies. An overview of the pipeline is provided in the Figure~\ref{fig:overall}.  
 
The pipeline consists of the following components:

\begin{itemize}
     \item \textbf{Action Space:} In our problem, the action space consists of payment policies, which are continuous variables. For non-continuous evaluators, we discretize them by grouping them into bins.

    \item \textbf{Policy Modeling:} Instead of directly using the original payment policy, we identified proxies by training Random Forest Classifiers (for discrete action space) / Regressors (for continuous action space) on the collected data. This approach is both generalizable and data-driven, making it applicable to any test scenario, regardless of the underlying model.

    \item \textbf{Reward Modeling:} A reward model is required for some of the metrics outlined in Section \ref{sec:Evaluators}. In these cases, we employed a Random Forest Regressor to model the reward.

    \item \textbf{Policy Evaluation:} Finally, we compare the rewards predicted by the OPE methods with the actual rewards observed during the  A/B tests to assess the accuracy of the estimators. Our experiments involve comparing a wide range of discrete and continuous evaluators. The evaluators used in these comparisons are listed in Section~\ref{sec:Evaluators}. The OPEs consist of some tunable hyperparameters which are learned to correct the distribution mismatch between the pre-tested policy and the new candidate policy. The data collected during the past A/B tests is used for this task.

    \item \textbf{Off-Policy Learning:} Off-policy learning is the approach where an agent learns about an optimal policy based on the off-policy evaluator.  Due to the dynamic nature of the policies and the complex underlying probability associated with them, we learn proxy payment models for simulation. We elaborate more about the process in Section~\ref{sec:off_policy_learning}. 
\end{itemize}

\subsection{Off-Policy Evaluation}
\label{sec:Evaluators}

The main idea behind using OPE in dynamic environments is to hypothetically test the performance of different payment policies offline before selecting one for the final A/B test. This additional evaluation allows us to identify the most promising payment policies from various options, thereby reducing the need for multiple A/B tests and lowering the chances of inconclusive results.

Various OPEs have been developed to enhance the accuracy of evaluations and manage noise in the data. OPEs can be categorized into two types based on the nature of the policy output: discrete and continuous. Discrete OPEs use the context-action distributions of the policies to generate importance weights for the data, which are then used to produce the final estimates. In contrast, continuous OPEs employ kernel density estimation to translate differences between policy actions into sample weights. Below, we provide a brief description of some of the evaluators considered in our work. A detailed description is available in the appendix.
 
\begin{itemize}

\item \textbf{Inverse Probability Weighting (IPW):~\cite{strehl2010learning, dudik2014doubly, su2020doubly}} Re-weights observed rewards by the ratio of the evaluation policy to the behavior policy, providing unbiased estimates but often with high variance.

\sloppy
\item \textbf{Self-Normalized Inverse Probability Weighting (SNIPW):~\cite{swaminathan2015selfnormalized, kallus2019intrinsically}} 
Normalizes rewards using self-normalized importance weights, trading off unbiasedness for increased stability.

\item \textbf{Direct Method (DM):~\cite{dudik2014doubly, beygelzimer2009offset}  } Estimates the policy value using a model of the expected rewards based on observed data, which depends heavily on the accuracy of the reward model.

\item \textbf{Doubly Robust (DR):~\cite{dudik2014doubly,farajtabar2018more, su2019cab,su2020doubly}} Combines IPW with a reward model to reduce variance, providing consistent estimates if either the importance weights or the reward model are accurate.

\item \textbf{Self-Normalized Doubly Robust (SNDR):~\cite{dudik2014doubly,kallus2019intrinsically}} Applies self-normalized importance weighting within the DR framework to improve stability while maintaining double robustness.

\item \textbf{Continuous Evaluators:~\cite{kallus2018policy}} Use kernel density estimation to model continuous treatment spaces, allowing for more precise OPE by avoiding information loss due to binning.

\end{itemize}

\subsubsection{Logged Data Collection}

 We motivate our approach by using three prior conducted A/B tests. We consider three A/B tests as listed below:

\begin{table}[h!]
\centering
\begin{tabular}{|c|c|c|}
\hline
\textbf{A/B Test} & \textbf{Control Policy} & \textbf{Treatment Policy} \\ 
\hline
Test-1 & \textbf{X} & \textbf{Y} \\ 
\hline
Test-2 & \textbf{X} & \textbf{Z} \\ 
\hline
Test-3 & \textbf{Y} & \textbf{Z} \\ 
\hline
\end{tabular}
\caption{Comparison of Policies in Two A/B Tests}
\label{table:Example}
\end{table}

In Table~\ref{table:Example}, we compare policies \textbf{X} and \textbf{Y} in \textit{Test-1}, and \textbf{X} and \textbf{Z} in \textit{Test-2}. Thus a major research question is -- ``\textit{Given the data from \textit{Test 1} and \textit{Test 2},  can we compare \textbf{Y} vs \textbf{Z} directly without conducting an actual A/B test?}''. To answer this question, we evaluate OPE approaches.  We use the results from the \textit{Test-3} as a ground truth for this comparison where we conducted an actual test between policies \textbf{Y} and \textbf{Z}.

The test measures are quantified by \textit{Lifts}, which are measured between the metrics in the treatment group and the control group\footnote{To ensure that the observed lift is not due to random chance, we perform a statistical significance test (T-Test) to get confidence intervals of the measured lifts.}:
\begin{flalign*}
    \text{Lift (\%)} = \left(\frac{\text{Metric}_{\text{Treatment}} - \text{Metric}_{\text{Control}}}{\text{Metric}_{\text{Control}}}\right) \times 100
\end{flalign*}

\vspace{-0.2cm}
\begin{figure}[htbp]
\centerline{\includegraphics[width=0.45\textwidth]{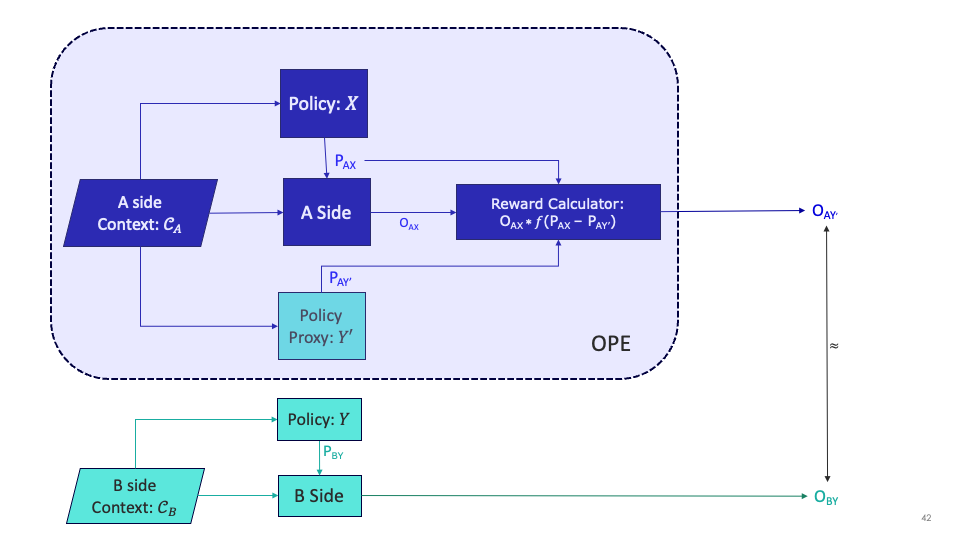}}
\caption{Off-Policy Evaluation Process Flow}
\label{fig:Flow}
\end{figure}
The flowchart in Fig~\ref{fig:Flow} illustrates the OPE process, where each test has two sides: Control (\textit{A-side}) and Treatment (\textit{B-side}). On the \textit{A-side}, context \(C_A\) is input to Policy \(X\), generating an reward outcome \(O_{AX}\) and predicted action \(P_{AX}\). A reward calculator then adjusts the reward outcome to \(O_{AY'}\) using a function \(f\) that compares \(P_{AX}\) with the action predicted by proxy policy \(Y'\), \(P_{AY'}\). On the \textit{B-side}, context \(C_B\) is used by Policy \(Y\), resulting in a reward outcome \(O_{BY}\) and predicted action \(P_{BY}\). The goal is to verify that \(O_{AY'}\), the OPE-derived reward outcome on \textit{A-side}  using \(Y'\), approximates \(O_{BY}\), the observed reward outcome on \textit{B-side}, assuming equivalent contexts \(C_A\) and \(C_B\). This comparison validates the OPE method's accuracy in predicting how Policy \(Y\) would perform on \textit{A-side}.

\subsubsection{Learning Policy Proxies}
Instead of using the policies directly for the evaluation, we learned proxies for the policies using random forest regression/classification models. Basically, we learned mappings from context space ($\mathcal{C}$) to the action space ($\mathcal{A}$) as: $\textbf{X}^{'}, \textbf{Y}^{'}, \textbf{Z}^{'}$  ($\pi : \mathcal{C} \rightarrow \mathcal{A}$). In case of discrete OPEs, $\mathcal{A}$ becomes a discrete space, while for continuous OPEs, it becomes a continuous space. For discrete case, a random forest classifier serves as the mapping structure, while a regressor is considered for the continuous case. For discrete scenarios, the continuous action space is divided into multiple bins and the actions are mapped to a specific bin. The bin numbers serve as the label for the data to be learned for the mapping.

 \begin{algorithm}[htbp]
\small\caption{  \textbf{Opt}imal \textbf{Pa}yment Policy \textbf{L}earning  (\texttt{OptPaL}) through OPE}
\begin{algorithmic}[1]
\Require Context data $\mathcal{C}$, Off-policy eval \texttt{OPE}, Learning rate $\alpha$, Max iterations $n_{max}$
\State Initialize \texttt{OptPaL} as an Multi-layer Perceptron (MLP) model with random weights $W$
\For{$i = 1$ to $n_{max}$}
    \State Predict payment using \texttt{OptPaL}: $\mathbf{P} \gets \texttt{OptPaL}(\mathcal{C})$
    \State Evaluate $\mathbf{P}$: $(Cost, Returns) \gets \texttt{OPE}(\mathcal{C}, \mathbf{P})$
    \State Compute loss: $\mathcal{L} \gets f(Cost, Returns)$ 
    \State Update weights: $W \gets W - \alpha \cdot \nabla \mathcal{L}$
    \If{Converged($\mathcal{L}$)}
        \State \textbf{break}
    \EndIf
\EndFor
\State \Return \texttt{OptPaL}
\end{algorithmic}
\label{alg:OPTRAL}
\end{algorithm}
 
\normalsize
 \subsection{New Policy Learning}
\label{sec:off_policy_learning}
Off-policy learning goes beyond mere evaluation by leveraging OPE results to iteratively refine and enhance policies. The primary objective of the new policy learning process is to identify a counterfactually optimal policy that maximizes expected rewards within the given environment for the past logged data. A key advantage of continuous off-policy evaluation is its differentiability, which facilitates the application of gradient descent methods to fit models and learn optimal policies. Following hyperparameter optimization, for continuous evaluators, we obtain individual OPEs for each of the outcome metrics under consideration.

Once the OPEs are established, we can employ gradient descent search on an underlying machine learning model to find the optimal payment policy based on a loss function involving \textit{Cost} and \textit{Returns}. The purpose of this exercise is to evaluate, hypothetically, how different payment strategies could have improved outcome metrics in the past. If we can develop a reliable proxy model for determining optimal payment policy, this model can be applied to inform future payment strategies or create the best-policy in the hindsight for comparison. We provide a description of the optimal policy learning using the off-policy evaluators in Algorithm~\ref{alg:OPTRAL}.

\section{Experimental Evaluation}

In our initial study, we pose two different research questions:

\begin{itemize}
    \item  \textit{Given two separate A/B Tests, conducted under different conditions, can we determine the best policy? }

    We consider two prior A/B tests as the basis of our evaluation which we call \textit{Test-1} and \textit{Test-2}. The results, as evaluated in the two tests, are provided in Fig~\ref{fig:ABTest}. At first glance, it seems that the \textbf{Policy Z} had the better lift, but    \textit{Test-1} and \textit{Test-2} are conducted under separate conditions. Thus it is difficult to say whether \textbf{Z} is truly better than \textbf{Y}. Thus our initial goal is to simulate under similar conditions which one is better in a simulated scenario of \textbf{Y} vs \textbf{Z}.

 \begin{figure}[htbp]
\centerline{\includegraphics[width=0.45\textwidth]{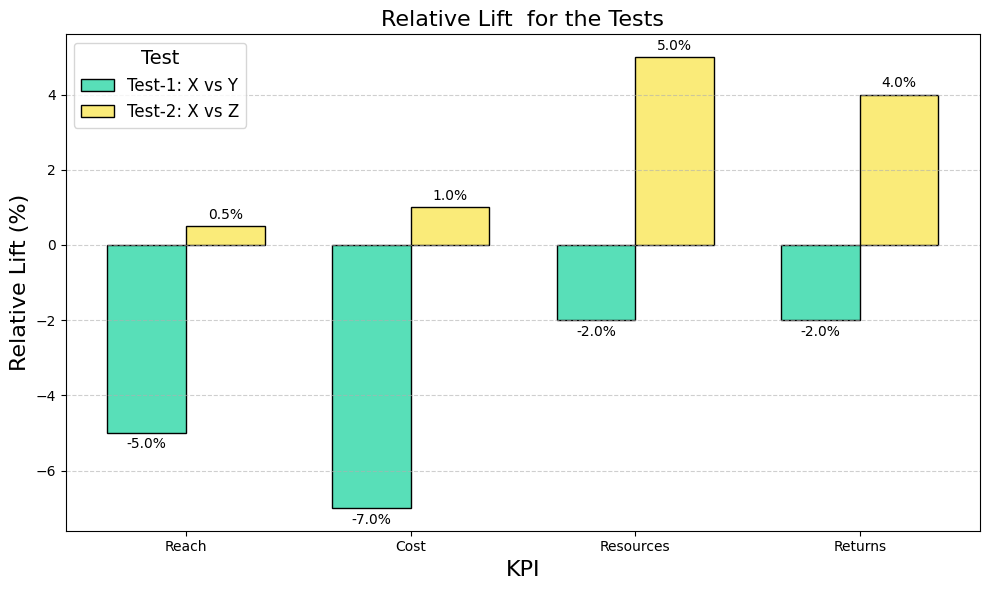}}
\caption{Relative Lifts for the two past A/B tests (\textit{Test-1} and \textit{Test-2}).}
\label{fig:ABTest}
\end{figure} 
 
\item  \textit{Can we learn an optimal policy from the data itself ?}

After assessing the suitability of OPE, we sought to explore counterfactual scenarios to determine what the optimal payment policy might have been in the past. Based on this analysis, we aim to learn an optimal policy by optimizing over the evaluation metrics.

\end{itemize}

 \subsection{Comparing Discrete vs Continuous case for payment}

\begin{figure}[t!]
\centering
\begin{subfigure}{.24\textwidth} 
\includegraphics[width=\textwidth]{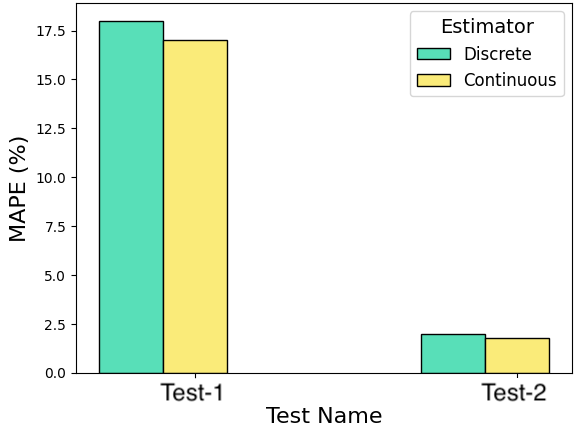}
\subcaption{Cost}
\label{fig:mape_2}
\end{subfigure}  
\begin{subfigure}{.24\textwidth}
\includegraphics[width=\textwidth]{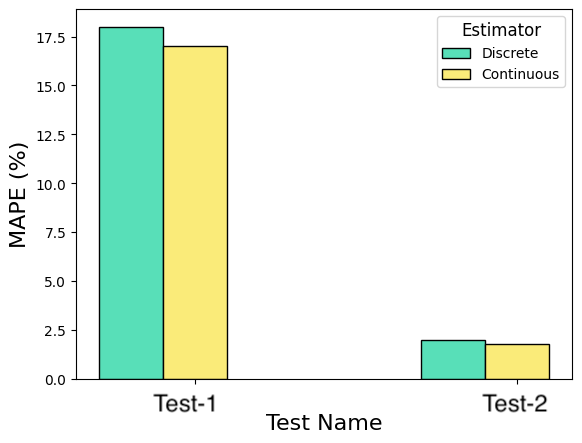} 
\subcaption{Reach}
\label{fig:mape_1}
\end{subfigure} 
\begin{subfigure}{.24\textwidth} 
\includegraphics[width=\textwidth]{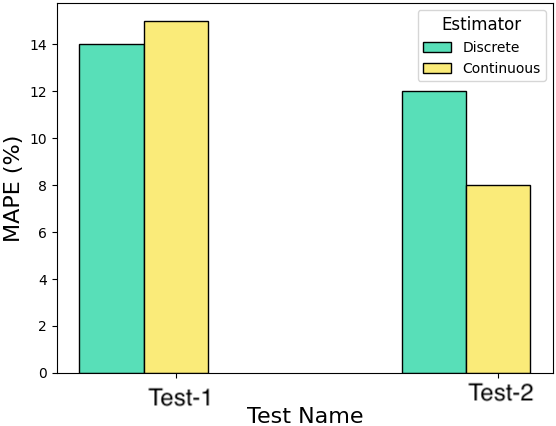}
\subcaption{Resources}
 \label{fig:mape_3}
\end{subfigure} 
\begin{subfigure}{.24\textwidth} 
\includegraphics[width=\textwidth]{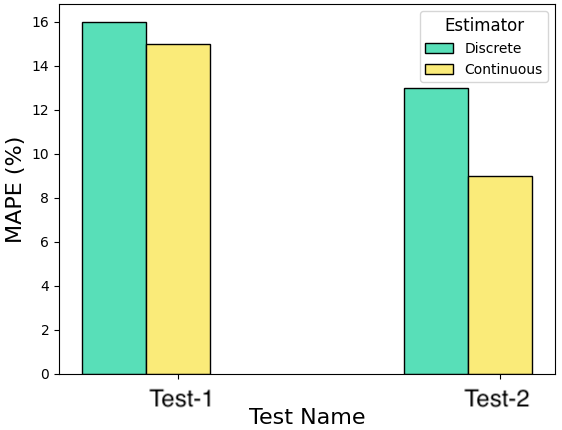}
\subcaption{Returns}
 \label{fig:mape_4}
\end{subfigure} 
\caption{MAPE Estimate: Discretized Action vs Continuous Action Evaluation }
\label{fig:estimator_comp}
\end{figure}

We need to compare the discretized action space and continuous action space to check how much accuracy we are gaining in terms of different metrics for moving to the continuous version. Both approaches are compared with respect to the Mean Absolute Percentage Error (MAPE) \cite{guha2023virtual}  for different metrics in the \textit{Test-1} and \textit{Test-2} experiments. The MAPEs aggregated over the OPE estimators for both tests are shown in Figure~\ref{fig:estimator_comp}.

\textbf{Observation-- \textit{Continuous OPE tends to outperform discretized OPE.}} \textit{This is intuitive because continuous OPE can capture the smooth, nuanced dependencies of the key metrics across small changes in payment values, providing a more detailed and accurate evaluation of policies. In contrast, discretized OPE may miss these fine-grained variations due to its coarser resolution. Overall, we get around 20\% reduction in MAPE by moving from the discretized version to the continuous version.}

 \begin{figure}[htbp]
\centering
\begin{subfigure}{.24\textwidth}
\includegraphics[width=\textwidth]{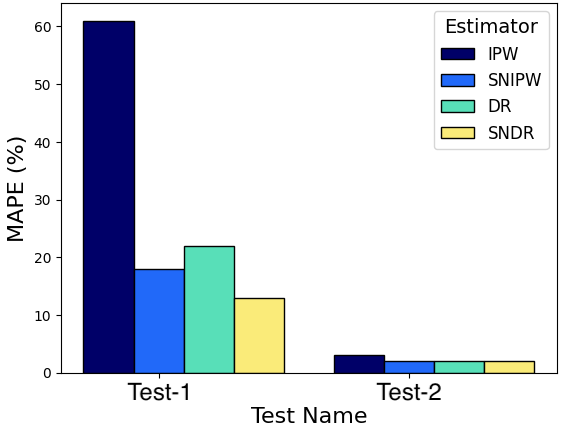}
\subcaption{Cost}
\label{fig:est_1}
\end{subfigure} 
\begin{subfigure}{.24\textwidth} 
\includegraphics[width=\textwidth]{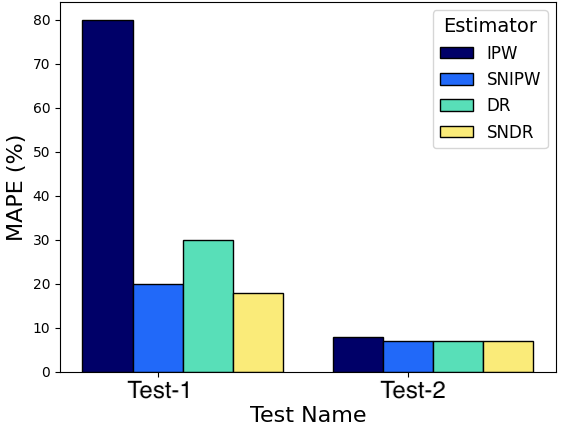}
\subcaption{Reach}
\label{fig:est_2}
\end{subfigure}  
\begin{subfigure}{.24\textwidth} 
\includegraphics[width=\textwidth]{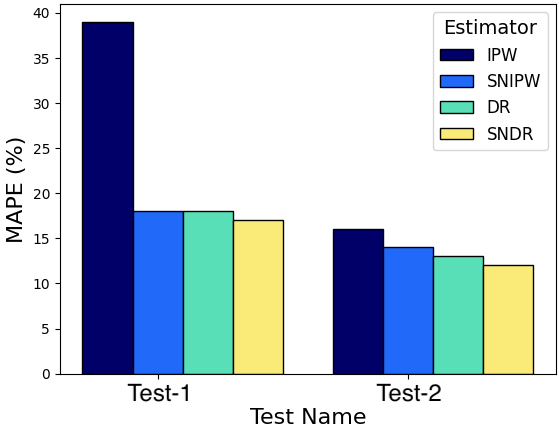}
\subcaption{Resources}
 \label{fig:est_3}
\end{subfigure} 
\begin{subfigure}{.24\textwidth} 
\includegraphics[width=\textwidth]{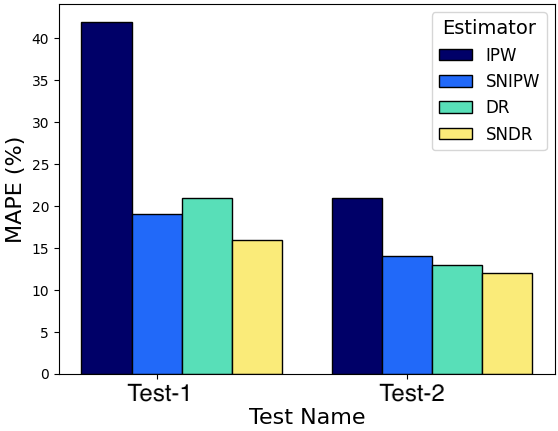}
\subcaption{Returns}
 \label{fig:est_4}
\end{subfigure} 
\caption{MAPE Estimate: Evaluation of Different Continuous   Evaluators }
\label{fig:discVcont}
\end{figure}
 \subsection{Comparison of Continuous Version of the Estimators}

 After verifying that continuous evaluators work better than the discrete evaluators, we evaluate the continuous version of the estimators mentioned in Section~\ref{sec:Evaluators}. In Figure~\ref{fig:discVcont}, we show the results of the top four evaluators for both tests.
 
\textbf{Observation-- \textit{SNDR works the best in our case.}} \textit{   The optimal kernel and bandwidth for the estimator can be determined through a separate hyperparameter tuning experiment. As more data is aggregated and used to predict overall rewards, the uncertainty in these predictions diminishes. For instance, predicting the total number of \textit{resources} acquired over a week is more reliable than predicting  at a daily grain. This reduction in uncertainty with larger aggregated data highlights the importance of selecting appropriate hyperparameters to achieve accurate predictions.
 }

 \subsection{Counterfactual  Test of Policies: \textbf{Y} vs   \textbf{Z}}

\subsubsection{Use Case I: Assessing the Accuracy of OPE}

\begin{figure}[htbp]
\centerline{\includegraphics[width=0.45\textwidth]{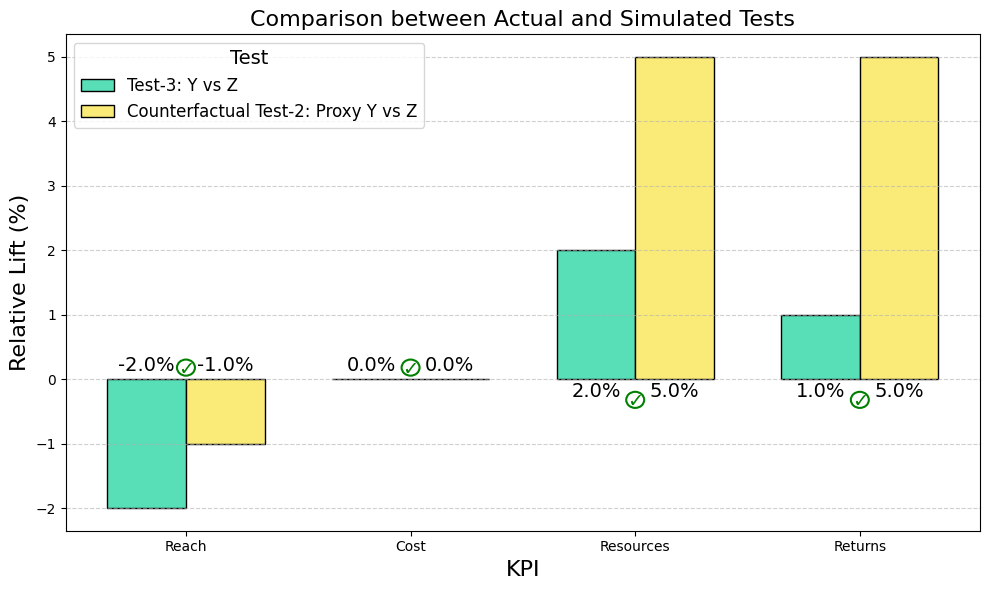}}
\caption{Relative Lifts for the Actual (\textbf{Y vs Z}) and Simulated (\textbf{Proxy Y vs Z}) Tests}
\label{fig:Counter_C_ABTest}
\end{figure}

To assess the accuracy of the learning proxies and the OPE, we conducted an evaluation — the \textit{Test-3}, comparing policies \textbf{Y} and \textbf{Z}. Following this, we constructed a \textbf{Proxy Policy Y} based on the past data collected for Policy Y. We then replaced \textbf{Proxy Policy Y} with Policy X in \textit{Test-2} which we refer to as \textit{Counterfactual Test-2} and compared the outcomes with the original \textit{Test-3} results, as illustrated in Figure~\ref{fig:Counter_C_ABTest}.

\indent\textbf{  Takeaways -- \textit{Directional Lifts are Correlated:}} \textit{For the mentioned metrics we were able to achieve similar directional lifts which shows that the OPEs were able to estimate the proper directional lifts for \textbf{Policy Y} through $\textbf{Proxy Policy Y}^{'}$. Improving the learning algorithms can potentially lead to accurate reward prediction and thus effective policy evaluation.}

\subsubsection{Use Case II: Estimating the Impact of \textbf{Policy Z} in  Test-1  in Place of \textbf{Policy X}, and Comparison with Counterfactual Test-2}

\begin{figure}[htbp]
\centerline{\includegraphics[width=0.45\textwidth]{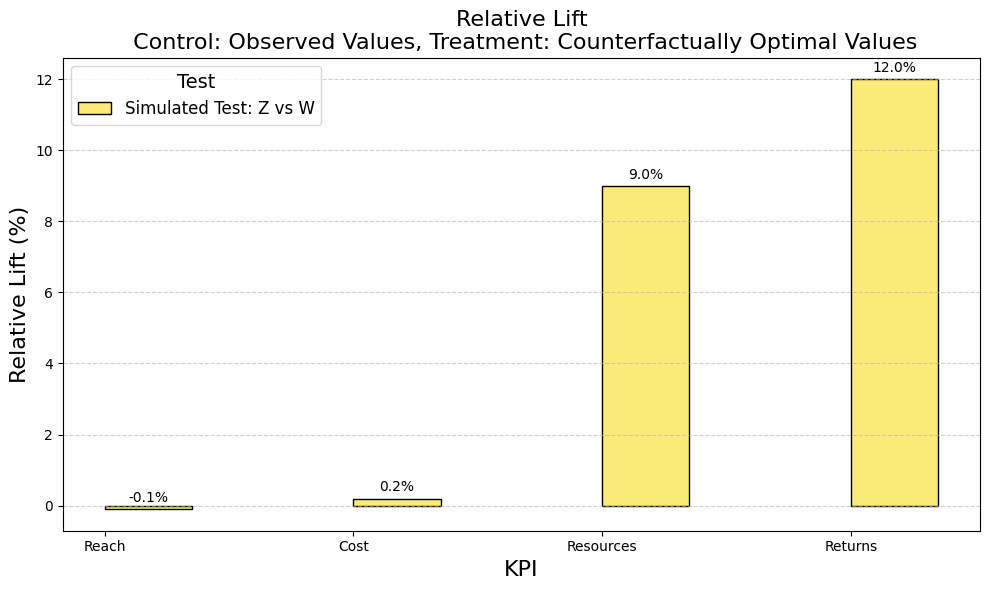}}
\caption{Relative Lifts for two Counterfactual Tests }
\label{fig:Counter_ABTest}
\end{figure}

After verifying that the OPEs are able to accurately predict the directional lifts for a counterfactual test, our next objective is to utilize the counterfactual evaluation process to compare different payment policies in different contexts without doing actual A/B tests. In the \textit{Test-1}, \textbf{Policy Y} showed negative lifts in \textit{reach}, \textit{resources}, and returns, while in the \textit{Test-2}, \textbf{Policy Z} demonstrated significant lifts in \textit{resources} acquired and the subsequent \textit{returns}. These results suggest that \textbf{Policy Z} is likely to outperform \textbf{Policy Y} in a direct comparison.\\
\indent To create a hypothetical evaluation, named as \textit{Counterfactual Test-1}, we simulated a scenario where a proxy version of \textbf{Policy Z} ($\textbf{Z}^{'}$) replaced \textbf{Policy X} as the control in the \textit{Test-1}. The treatment was set to \textbf{Policy Y}.
The resulting lifts for \textit{Counterfactual Test-1} and \textit{Counterfactual Test-2} scenarios are compared in Fig~\ref{fig:Counter_ABTest}. In \textit{Counterfactual Test-1}, the lifts indicate that \textbf{Proxy Policy Z} outperforms \textbf{Policy Y} across all KPIs. In \textit{Counterfactual Test-2}, although \textbf{Policy Z} shows a slight negative lift in \textit{Reach}, this is offset by substantial positive lifts in \textit{Resources} and \textit{Returns}.\\
\indent\textbf{  Insights -- \textit{Robustness of Policy Z:}} \textit{ The consistency in performance of \textbf{Policy Z} across both tests suggests that it is generally a more robust choice for improving key metrics.}

 \subsection{New Policy Learning through OPE Optimization}

Finally, we conducted new model learning to identify the optimal policy in retrospect. For our use case, we selected profit maximization as the objective function. As a standard practice in machine learning, the negative of the profit maximization is used as the loss function, defined as:
 
\vspace{-0.1cm}
\begin{flalign*}
    \min_{p} f &= - \left( \text{Returns}_{\texttt{OPE}}(p) - \text{Cost}_{\texttt{OPE}}(p) \right),\\
    &\text{where } p = \texttt{OptPaL}(\mathcal{C})
\end{flalign*}

Here, the returns and cost are estimated using the OPE models. We employed a multi-layer perception (MLP) named $\texttt{OptPaL}$ with two hidden layers having $\frac{|\mathcal{C}|}{2}$ and $\frac{|\mathcal{C}|}{4}$ nodes, respectively. The input to $\texttt{OptPaL}$ is the context $\mathcal{C}$, while the output is the payment value $p$. By optimizing $\texttt{OptPaL}$ through gradient descent, we achieved a non-diminishing objective value after training the counterfactual model for 1000 epochs. To compare the relative performance of the new optimal policy (\textbf{Policy W}: \texttt{OptPaL($\mathcal{C}$)}) with the existing \textbf{Policy X}, we utilized the \textit{Test-2}.

\begin{figure}[t]
\centerline{\includegraphics[width=0.45\textwidth]{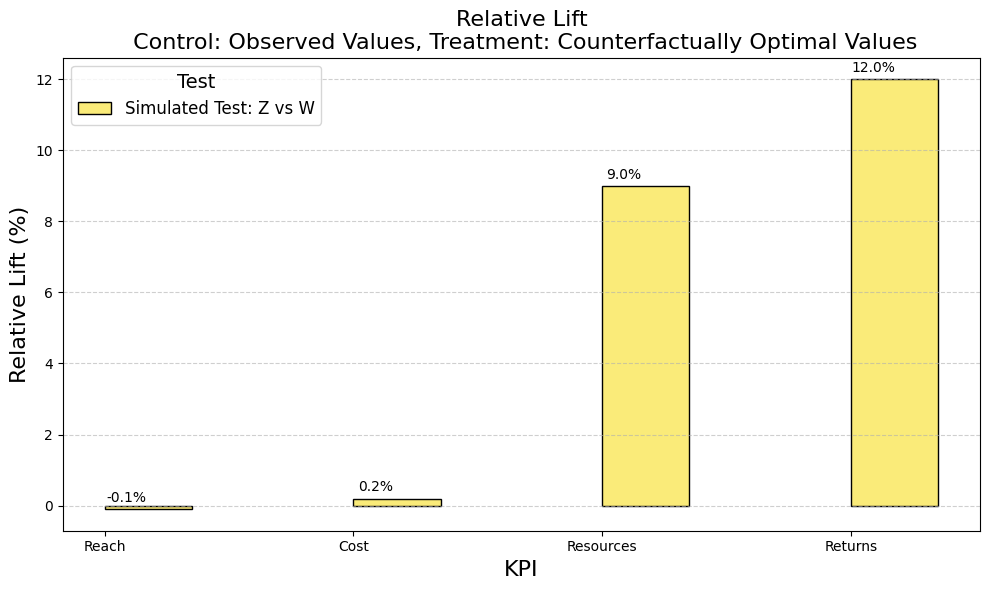}}
\caption{Relative Lifts for the Actual and Simulated Tests}
\label{fig:Counter_sim_ABTest}
\end{figure}

The results of the simulated test are shown in Figure~\ref{fig:Counter_sim_ABTest}. The optimal policy \textbf{W} was able to allocate the same expenditure as \textbf{Policy X} while maintaining comparable \textit{Reach} and \textit{Cost}, but it increased the overall \textit{Resource} count and total \textit{Return}. This demonstrates the efficacy of \textbf{Policy W} and provides insights into the predictive uncertainty by showing how much error was made due to the limitations in our predictions.

\textbf{Takeaways:} \textit{Counterfactual policy learning helps us discover new optimal policies and can be used for baseline evaluations of new policies.}

 \section{Conclusion}

In this study, we have demonstrated the effectiveness of off-policy evaluation (OPE) techniques in refining payment strategies by leveraging counterfactual scenarios. Our results underscore the advantages of continuous OPE methods over traditional discretized approaches, providing more precise and nuanced insights into policy performance. Through counterfactual policy evaluation, we are able to estimate directional lifts for new policies, offering a preliminary assessment of their utility prior to conducting A/B tests. Furthermore, we developed optimal policies by optimizing the continuous estimator, thereby identifying the most effective strategies in retrospect.

\section{Future Vision}
\begin{figure}[htbp]
\centerline{\includegraphics[width=0.45\textwidth]{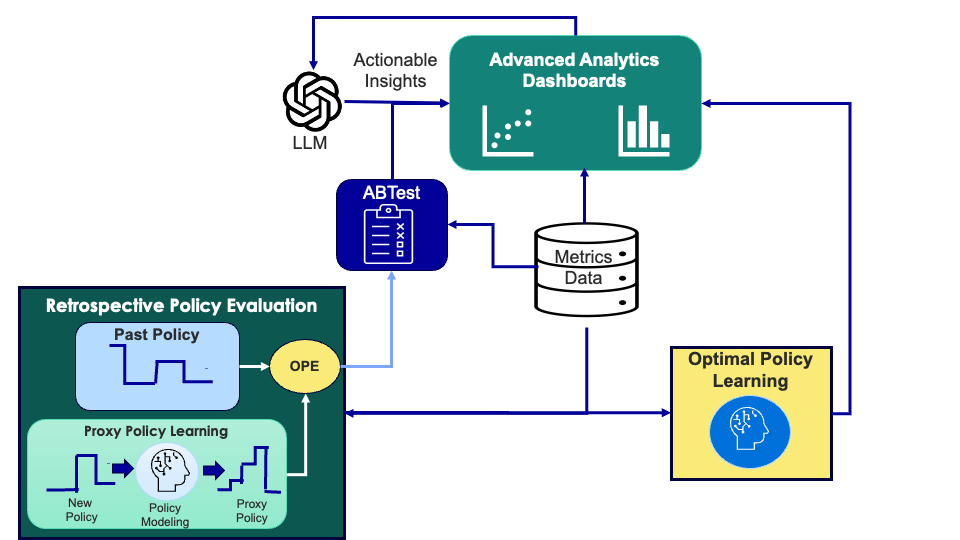}}
\caption{Advanced Analytics Platform}
\label{fig:analytics}
\end{figure}

The future of policy evaluation envisions a fully automated system designed to seamlessly and dynamically assess new payment strategies and policies. This system will integrate scalable engineering and advanced methodologies to deliver continuous, accurate, and actionable insights. We provide a description of the system design of the Advanced Analytics Platform in the Figure~\ref{fig:analytics}.

The foundational components of the system will include:

\begin{itemize}
    \item \textbf{Retrospective Policy Evaluation:} This component focuses on learning the optimal policy in hindsight, allowing us to determine the best course of action based on past data. It involves simulating outcomes of new algorithmic policies, offering foresight into potential impacts and enabling the evaluation of policies before deployment. This subsystem will help decide whether a new proposed policy should undergo A/B testing.
    
    \item \textbf{Optimal Policy Learning:} Aims to generate an  optimal policy from historical data. This will help us compare and   analyze the regret of past actions.  
    
    \item \textbf{Comprehensive Monitoring and Insights:} The system will provide intuitive and actionable insights into policy performance, trends, and recommendations, enabling decision-makers to understand and respond effectively. Integrating Large-Language Models will further refine the learning process and enhance actionable insights.
\end{itemize}

Our next research direction will focus on enhancing proxy policy learning and exploring Structural Causal Model-Based Counterfactual Evaluation~\cite{balke2022probabilistic} to improve the accuracy of policy assessments. Additionally, we will consider large dimensional action spaces~\cite{saito2022off} to better account for complex auction environments.

\appendix

\subsection{Evaluators Detailed Description}

\subsubsection{Inverse Probability Weighting (IPW)}
IPW estimates the policy value of evaluation policy($\pi_e$)~\cite{strehl2010learning, dudik2014doubly, su2020doubly}:  

\begin{equation}
    \hat{V}_{\mathrm{IPW}} (\pi_e; \mathcal{D}) := \mathbb{E}_{n} [ w(x_i,a_i) \cdot r_i]
\end{equation}

where  $\mathcal{D}=\{(x_i,a_i,r_i)\}_{i=1}^{n}$ is logged bandit data with  n observations collected by behavior policy $\pi_b$; $w(x,a):=\pi_e (a|x)/\pi_b (a|x)$   is the importance weight given  $x$  and $a$. $\mathbb{E}_{n}[\cdot]$  is the empirical average over  $n$  observations in $D$ . When the clipping is applied, a large importance weight is clipped as  $\hat{w}(x,a) := \min \{ \lambda, w(x,a) \}$ where  $\lambda(>0)$ is a hyperparameter to specify a maximum allowed importance weight.
IPW re-weights the rewards by the ratio of the evaluation policy and behavior policy (importance weight). When the behavior policy is known, IPW is unbiased and consistent for the true policy value. However, it can have a large variance, especially when the evaluation policy significantly deviates from the behavior policy.

\subsubsection{Self-Normalized Inverse Probability Weighting (SNIPW)}

SNIPW estimates the policy value of evaluation policy ($\pi_e$)~\cite{swaminathan2015selfnormalized, kallus2019intrinsically} as \begin{equation}
    \hat{V}_{\mathrm{SNIPW}} (\pi_e; \mathcal{D}) := \frac{\mathbb{E}_{n} [w(x_i,a_i) \cdot r_i]}{ \mathbb{E}_{n} [w(x_i,a_i)]}
\end{equation} 

SNIPW normalizes the observed rewards by the self-normalized importance weight. This estimator is not unbiased even when the behavior policy is known. However, it is still consistent for the true policy value and gains some stability in OPE.

\subsubsection{Direct Method (DM)}

DM~\cite{dudik2014doubly, beygelzimer2009offset} first trains a supervised ML model, such as ridge regression and gradient boosting, to estimate the reward function $q(x,a) = \mathbb{E}[r|x,a]$. It then uses the estimated rewards to estimate the policy value as follows.

    \begin{align}
\hat{V}_{\mathrm{DM}} (\pi_e; \mathcal{D}, \hat{q})
        &:= \mathbb{E}_{n} \left[ \sum_{a \in \mathcal{A}} \hat{q} (x_i,a) \pi_e(a|x_i) \right]   \nonumber  \\
        & =  \mathbb{E}_{n}[\hat{q} (x_i,\pi_e)] 
\end{align}
 
 $\hat{q} (x,a)$  is the estimated expected reward given  \textit{x} and \textit{a}. $\hat{q} (x_i,\pi):= \mathbb{E}_{a \sim \pi(a|x)}[\hat{q}(x,a)]$   is the expectation of the estimated reward function over $\pi$ . If the regression model ($\hat{q}$) is a good approximation to the true mean reward function, this estimator accurately estimates the policy value of the evaluation policy. If the regression function fails to approximate the reward function well, however, the final estimator is no longer consistent.

\subsubsection{Doubly Robust (DR)}

Similar to DM, DR~\cite{dudik2014doubly,farajtabar2018more, su2019cab,su2020doubly},  estimates the reward function ( $q(x,a)=\mathbb{E}[r|x,a]$  ). It then uses the estimated rewards to estimate the policy value as follows.

\begin{align}
    \hat{V}_{\mathrm{DR}} (\pi_e; \mathcal{D}, \hat{q})
        := \mathbb{E}_{n}[\hat{q}(x_i,\pi_e) +  w(x_i,a_i) (r_i - \hat{q}(x_i,a_i)]
\end{align}

When the clipping is applied, a large importance weight is clipped as $\hat{w}(x,a) := \min \{ \lambda, w(x,a) \}$ where  $\lambda(>0)$ is a hyperparameter to specify a maximum allowed importance weight.
DR mimics IPW to use a weighted version of rewards, but DR also uses the estimated mean reward function (the regression model) as a control variate to decrease the variance. It preserves the consistency of IPW if either the importance weight or the mean reward estimator is accurate (a property called double robustness). Moreover, DR is semi-parametric efficient when the mean reward estimator is correctly specified.

\subsubsection{Self-Normalized Doubly Robust (SNDR)}

Similar to SNIPW, the SNDR estimator applies the self-normalized importance weighting technique to gain some stability. The SNDR estimator computes the policy value of the evaluation policy $\pi_e$ as
\begin{flalign}
    \hat{V}_{\mathrm{SNDR}} (\pi_e; \mathcal{D}, \hat{q}) :=
        \mathbb{E}_{n} \left[\hat{q}(x_i,\pi_e) +  \frac{w(x_i,a_i) (r_i - \hat{q}(x_i,a_i))}{\mathbb{E}_{n}[ w(x_i,a_i) ]} \right]
\end{flalign}  

\subsubsection{Continuous Evaluators}

In traditional discretized off-policy evaluators, significant drawbacks arise due to the loss of information when continuous values are binned together. This binning process can obscure subtle differences within the data, making it challenging to evaluate small changes, as these nuances are often lost within the bins. Furthermore, discretized approaches rely on ad-hoc modeling of the policies, which may not accurately reflect the underlying continuous nature of the data.

In contrast, continuous off-policy evaluation (OPE) is more appropriate for contexts with continuous treatment spaces. To address these challenges, we employ Multivariate Kernel Density Estimation (KDE) to model the conditional probability density of the continuous variables (e.g., payment, denoted as \( b \)) with respect to the context (denoted as \( x \)):

\begin{flalign*}
    P(T=b|X=x) = \frac{P(T=b,X=x)}{P(X=x)}
\end{flalign*}

Once we model the probability densities, different kernels are applied to map the differences between payments to importance weights. Essentially, the OPE strategy assigns greater weight to data samples that exhibit smaller differences in payment amounts relative to the currently evaluated sample. The estimated reward is calculated using the continuous evaluation metric~\cite{kallus2018policy}, which is defined as:

\begin{flalign}
    \hat{v_\tau} = \frac{1}{nh} \sum_{i=1}^{n} K \left(\frac{\tau(x_i)-t_i}{h}\right) \frac{y_i}{Q_i}
\end{flalign}

In this equation:

\begin{itemize}
    \item $n$   is the number of data samples,
 \item    $h$  is the bandwidth,
 \item     $K$  is the kernel function,
 \item     $\tau(x_i)$  represents the action that would have been taken by the evaluation policy,
 \item    $t_i$  represents the action taken by the behavioral policy,
 \item    $Q_i = P(t_i\|x_i)$  is the probability density of the action under the behavioral policy, and
 \item   $y_i$ is the observed reward.
\end{itemize}

Two critical hyperparameters in this process are the kernel \( K \) and the bandwidth \( h \). The bandwidth controls the scale of proximity considered between treatments: a bandwidth too large introduces high bias as it averages over a broader dataset, while a bandwidth too small increases variance by focusing too narrowly. Therefore, careful tuning of both the kernel and bandwidth is essential to optimize the OPE process for specific key metrics.

To perform hyperparameter tuning, we utilize Optuna~\cite{optuna_2019}, an efficient optimization framework, to individually optimize the kernel and bandwidth for each metrics.

\vfill

\end{document}